\documentclass{article}

\usepackage{microtype}

\usepackage{graphicx}
\usepackage{subfigure}
\usepackage{booktabs} 
\usepackage[absolute,overlay]{textpos} 

\usepackage{hyperref}

\usepackage[accepted]{arxiv}

\usepackage{amsmath}
\usepackage{amssymb}
\usepackage{mathtools}
\usepackage{amsthm}

\usepackage[capitalize,noabbrev]{cleveref}

\theoremstyle{plain}

\theoremstyle{definition}

\theoremstyle{remark}

\usepackage[textsize=tiny]{todonotes}

\arxivtitlerunning{Towards Robust and Reliable Concept Representations: Reliability-Enhanced Concept Embedding Model}

\begin{document}

\twocolumn[
\arxivtitle{Towards Robust and Reliable Concept Representations:\\ Reliability-Enhanced Concept Embedding Model}



\arxivsetsymbol{equal}{*}

\begin{arxivauthorlist}
\arxivauthor{Yuxuan Cai}{ntu}
\arxivauthor{Xiyu Wang}{ntu}
\arxivauthor{Satoshi Tsutsui}{ntu}
\arxivauthor{Winnie Pang}{ntu}
\arxivauthor{Bihan Wen}{ntu}
\end{arxivauthorlist}

\arxivaffiliation{ntu}{Nanyang Technological University, Singapore}

\arxivcorrespondingauthor{Bihan Wen}{bihan.wen@ntu.edu.sg}

\vskip 0.3in
]



\printAffiliationsAndNotice{\arxivEqualContribution} 

\begin{abstract}

Concept Bottleneck Models (CBMs) aim to enhance interpretability by predicting human-understandable concepts as intermediates for decision-making. However, these models often face challenges in ensuring reliable concept representations, which can propagate to downstream tasks and undermine robustness, especially under distribution shifts. 
Two inherent issues contribute to concept unreliability: sensitivity to concept-irrelevant features (e.g., background variations) and lack of semantic consistency for the same concept across different samples. To address these limitations, we propose the Reliability-Enhanced Concept Embedding Model (RECEM), which introduces a two-fold strategy: Concept-Level Disentanglement to separate irrelevant features from concept-relevant information and a Concept Mixup mechanism to ensure semantic alignment across samples. These mechanisms work together to improve concept reliability, enabling the model to focus on meaningful object attributes and generate faithful concept representations. Experimental results demonstrate that RECEM consistently outperforms existing baselines across multiple datasets, showing superior performance under background and domain shifts. These findings highlight the effectiveness of disentanglement and alignment strategies in enhancing both reliability and robustness in CBMs.
\end{abstract}

\begin{figure}[!t]
    \centering
    \includegraphics[trim=0 378pt 0 0, clip, width=\columnwidth]{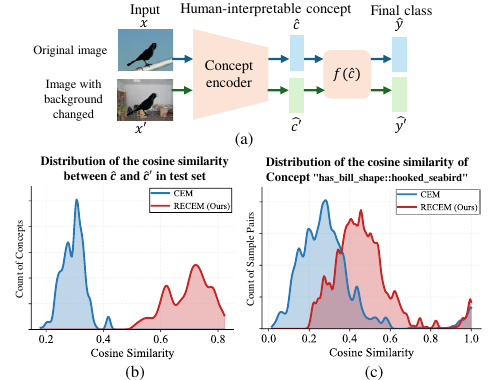}
    \caption{(a) The model predicts \(\hat{c}\) and \(\hat{c}'\), corresponding to the original image and the image with a background shift. Reliable concept representations should be robust to irrelevant variations and exhibit consistent semantic meanings. (b) shows the cosine similarity distribution between embeddings before and after the background change, reflecting model sensitivity to irrelevant features. The results indicate that CEM (blue) has lower similarity, reflecting a lack of consistency and sensitivity to irrelevant features. (c) presents the similarity distribution of the same concept representation across different samples in the CUB dataset, illustrating semantic consistency. It demonstrates that CEM (blue) exhibits poor semantic consistency across samples.}
    \label{fig:introfig}
\end{figure}

\section{Introduction}

With the increasing demand for reliable artificial intelligence, explanatory deep neural networks (DNNs) have attracted a lot of attention in various research fields. While DNNs are capable of achieving superior performance, their decision-making process is often opaque and therefore difficult to trust and verify. To address this, concept Bottleneck Models (CBMs)~\cite{koh2020concept, kim2023probabilistic, yuksekgonul2022post} have emerged as a promising approach to balance interpretability with performance. These models divide the decision-making process into two stages: first, predicting high-level, human-understandable concepts from raw input data, and second, using these predicted concepts to derive the final class-label predictions. CBMs offer an interpretability advantage by providing intermediate concepts that are comprehensible to humans, thus enabling insights into the model’s reasoning. Moreover, they allow human experts to make intervention during predictions~\cite{koh2020concept}, modifying final class outputs by adjusting the intermediate concepts.

Despite their significant potential, CBMs face challenges in ensuring reliable concept representations, which can propagate to downstream tasks and undermine robustness and reliability~\cite{furby2023towards, havasi2022addressing}. Inspired by  Concept Embedding Models (CEMs)~\cite{zarlenga2022concept}, we hypothesize that the unreliability of concept embeddings arises from two interrelated issues: sensitivity to irrelevant variations and lack of semantic consistency.
\begin{figure}[!t]
    \centering
    \includegraphics[trim=0 478pt 0 0, clip, width=\columnwidth]{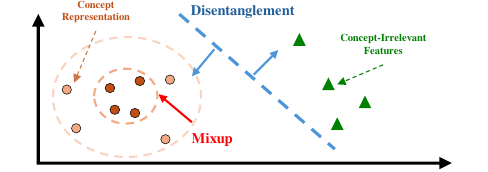}
    \caption{An intuitive illustration of the proposed mechanisms, Concept-Level Disentanglement and Concept Mixup.}
    \label{fig:introfig2}
\end{figure}

Sensitivity to concept-irrelevant features, such as background changes, is a significant challenge for CBMs. Ideally, predicted concepts should correspond to meaningful regions of the input and remain robust against irrelevant features. To evaluate this issue, we conduct an experiment using the TravelingBirds dataset, a variant of the CUB dataset with altered backgrounds. By measuring the cosine similarity between concept embeddings of original test images from CUB and their counterparts in TravelingBirds, we observe significant variability in the embeddings generated by existing CEMs. As shown in Figure~\ref{fig:introfig}, the distribution of similarity values for CEM reflects substantial inconsistency, highlighting their sensitivity to concept-irrelevant features and their inability to effectively disentangle concept-relevant information from irrelevant noise.

A related issue is the lack of semantic consistency for the same concept across different samples. Reliable concept embeddings should exhibit high semantic similarity when representing the same concept, reflecting consistent semantic representations. Inspired by~\cite{liu2021activity}, we analyze the cosine similarity distribution of a concept chosen as an example across samples in the CUB dataset. Figure~\ref{fig:introfig} reveals that the representations for this example concept often vary significantly between samples. This inconsistency undermines the reliability of predicted concepts and limits the interpretability of CBMs in practical applications.

To address these challenges, we propose two complementary mechanisms based on CEM~\cite{zarlenga2022concept}: Concept-Level Disentanglement and Concept Mixup. Together, these methods decouple concept-relevant features from irrelevant ones and align the semantic representations of the same concept across samples. Figure~\ref{fig:introfig2} provides an intuitive illustration of how these mechanisms improve semantic consistency and effectively separate concept-relevant and irrelevant features. This approach significantly improves the concept reliability, especially in scenarios where background noise or irrelevant features can affect the accuracy of concept predictions. Our contributions can be summarized as follows:

\begin{itemize}
    \item We analyze the limitations of CEMs, focusing on two critical challenges: the sensitivity of concept embeddings to irrelevant variations, such as background changes, which undermines robustness, and  semantic inconsistency across different samples for the same concept, which reduces reliability.
    \item We propose a novel concept embedding model architecture that integrates Concept-Level Disentanglement and Concept Mixup to improve concept reliability, as well as robustness and performance in real-world scenarios.
    \item We demonstrate that RECEM outperforms all baseline methods, especially when the background is shifted. Also, our proposed method has higher conceptual alignment than the baseline and has a stronger ability to accept human expert interventions.
\end{itemize}

\section{Related Work}

\textbf{Concept Bottleneck Models} \cite{koh2020concept, kim2023probabilistic, yuksekgonul2022post} have emerged as a promising approach to bridge the gap between interpretability and performance in deep learning. CBMs operate by first predicting high-level, human-understandable concepts, which are then used as intermediates to make final task predictions. 
Building upon CBMs, Concept Embedding Models \cite{zarlenga2022concept} extend the framework by introducing high-dimensional embeddings for concepts. By characterizing each concept with a pair of positive and negative embeddings, CEMs aim to represent the variability within concepts. Post-hoc Concept Bottleneck Models (PCBM) \cite{yuksekgonul2022post} employ a residual fitting mechanism to refine concept predictions, thereby enhancing task accuracy. Probabilistic Concept Bottleneck Models (ProbCBMs) \cite{kim2023probabilistic} incorporate uncertainty estimates into concept predictions, allowing the model to account for ambiguity in the input data. However, some recent work \cite{margeloiu2021concept} has pointed out that many of the learned concepts of CBMs do not learn exactly as intended, thus weakening the interpretability of CBMs, but current work is limited to visualising examples or requiring complete changes to the reasoning process of CBMs, and we urgently need an architecture and learning strategy that can effectively mitigate this problem in order to move CBMs forward.

\textbf{Disentangled Representation Learning} Disentangled Representation Learning (DRL) aims to learn representations that separate underlying factors of variation in data into independent, semantically meaningful variables. By doing so, DRL improves explainability, generalizability, and controllability across various tasks such as computer vision~\cite{cheng2023disentangled}, natural language processing~\cite{wu2020improving}, and graph learning~\cite{wang2024disentangling}. Common approaches include generative models like VAEs~\cite{xu2021multi}and GANs~\cite{liu2020oogan,tran2017disentangled}, which incorporate additional regularizers to enforce independence among latent factors, as well as methods based on causal inference and group theory~\cite{bengio2013representation, higgins2018towards}. DRL has been successfully applied in synthetic datasets for dimension-wise disentanglement~\cite{chen2016infogan,jeon2021ib} and real-world applications for vector-wise disentanglement~\cite{liu2021activity,cheng2023disentangled}, with the former focusing on fine-grained factors and the latter on coarse-grained ones. Despite significant progress, challenges remain, such as balancing disentanglement quality and task performance, as well as addressing dependencies among generative factors~\cite{suter2019robustly}.

\section{Preliminaries}

\textbf{Concept Bottleneck Models} \cite{koh2020concept, kim2023probabilistic} are designed to enhance the interpretability of deep neural networks by introducing an intermediate bottleneck layer composed of human-defined concepts. CBMs decompose the learning process into three components: a backbone network \(\psi\), a concept encoder \(g\), and a downstream predictor \(f\). Given a dataset \(\mathcal{D} = \{(x_i, c^{gt}_{i,k}, y_{i,m})\}_{i=1}^N\), where \(x_i \in \mathcal{X}\) is the input, \(c^{gt}_{i,k} \in \mathcal{C} = \{0, 1\}\) represents binary concept labels for \(K\) concepts, and \(y_{i,m} \in \mathcal{Y} \subset \{0, 1\}\) is the task label, CBMs predict \(y\) via intermediate concept representations \(\hat{C}\).

The backbone network \(\psi\) maps the input \(x\) to a latent representation \(h \in \mathbb{R}^{n_{\text{hidden}}}\), i.e., \(h = \psi(x)\). The concept encoder \(g\) predicts the concept vector \(\hat{c} = g(h)\), while the downstream predictor \(f\) maps \(\hat{c}\) to the final output \(\hat{y} = f(\hat{C})\). This modular structure enables interpretability by allowing human intervention at the concept level. 

\textbf{Concept Embedding Models} \cite{zarlenga2022concept} extend CBMs by introducing high-dimensional embeddings to represent concepts. For each concept \(c^{gt}_k\), two \(m\)-dimensional embeddings are learned: \(c_k^+ \in \mathbb{R}^d\) for the active state (\(c^{gt}_k = 1\)) and \(c_k^- \in \mathbb{R}^d\) for the inactive state (\(c^{gt}_k = 0\)). These embeddings are generated by a learnable model \(\phi_k^\pm: \mathbb{R}^{n_{\text{hidden}}} \to \mathbb{R}^d\), implemented as deep neural networks. A shared scoring function \(s: \mathbb{R}^{2d} \to [0, 1]\) is used to estimate the probability \(\hat{p}_k = s([\phi_k^+(h), \phi_k^-(h)])\), where \([\phi_k^+(h), \phi_k^-(h)]\) represents the concatenation of the two embeddings. 

The final concept representation \(\hat{c}_k\) is constructed as:
\begin{equation}
\hat{c}_k = \hat{p}_k \phi_k^+(h) + (1 - \hat{p}_k) \phi_k^-(h).
\end{equation}
The complete concept vector \(\hat{C} = [\hat{c}_1, \dots, \hat{c}_K] \in \mathbb{R}^{Kd}\) is then passed to a learnable label predictor \(f(\hat{C})\), which outputs the predicted label \(\hat{y}\). By leveraging high-dimensional embeddings, CEMs capture richer concept representations, making them more robust to complex tasks and incomplete concept annotations.

\section{Method}
In concept embedding models, semantic inconsistencies and unreliable conceptual representations are key challenges affecting downstream tasks. We analyze these key challenges and propose a unified framework, RECEM, to address them. Our approach focuses on ensuring that semantic representations of the same concept remain consistent across samples and categories, and improving the separation of conceptually relevant and irrelevant features to increase the reliability of concept embeddings. To achieve this goal, RECEM introduces two key components: the Concept-Level Disentanglement framework and the Concept Mixup mechanism. These two components work in tandem to align semantic representations and improve the robustness of the concept encoder. In this section, we describe each component and its integration in the overall framework in detail.
\begin{figure*}[!t] 
    \centering
    \includegraphics[trim=0 0 0 0, clip,width=\textwidth]{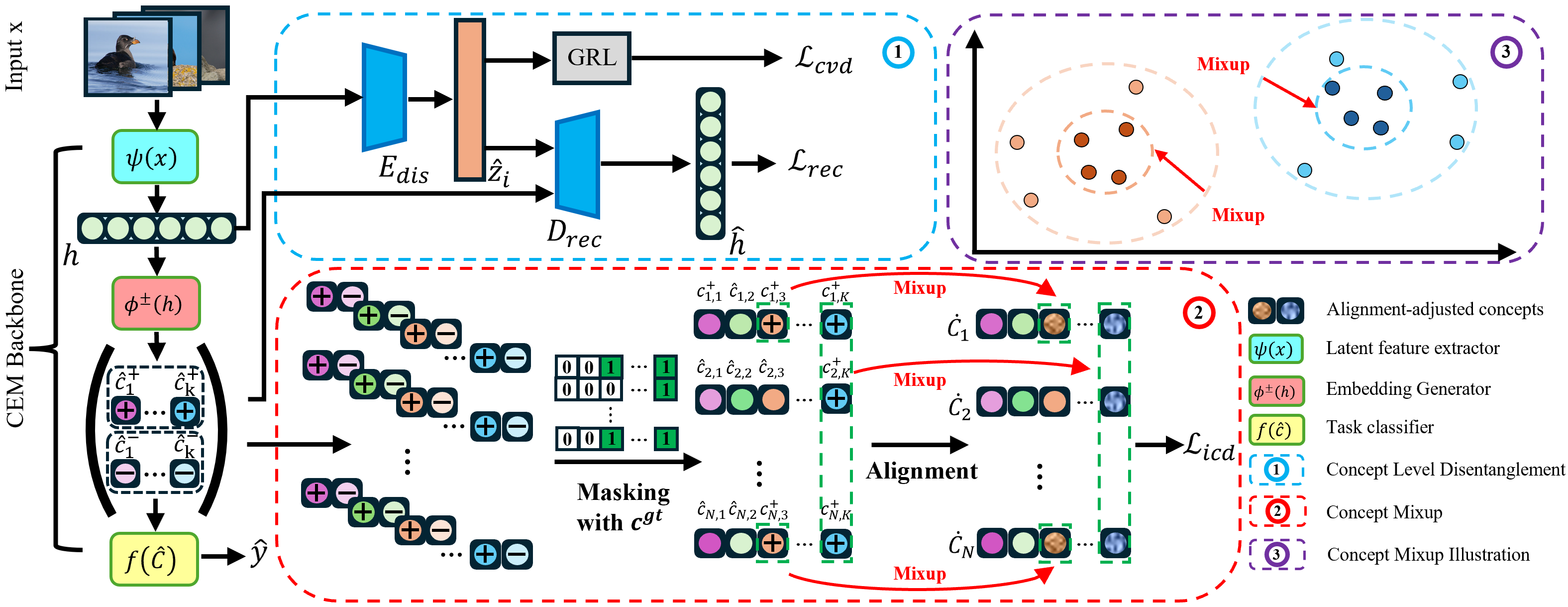}
    \caption{Overview of the proposed RECEM architecture, highlighting three key components with numbered dashed boxes: 
    {\color[HTML]{00B0F0} (1)} {\color[HTML]{00B0F0} Concept-Level Disentanglement}, where {\color[HTML]{0070C0} \(E_{\text{dis}}\)} extracts concept-irrelevant features {\color[HTML]{C0643E} \(\hat{z}_i\)} and {\color[HTML]{0070C0} \(D_{\text{rec}}\)} ensures rich semantic information within the concept embeddings; 
    {\color{red} (2)} {\color{red} Concept Mixup}, a mechanism aligning and proportionally mixing {\color[HTML]{00B050} true concept embeddings} (indicated by the {\color[HTML]{00B050} green dashed box}) across samples to achieve semantic consistency of concepts; 
    and {\color[HTML]{800080} (3)} {\color[HTML]{800080} Illustration of }{\color{red} Concept Mixup}, where arrows demonstrate the Concept Mixup mechanism enhances the consistency of semantic representations for the same concept after alignment.}

    \label{fig:model_architecture}
\end{figure*}

\subsection{Problem Definition}

We consider a supervised classification setting with \(N\) data points, \(K\) concepts, and \(M\) classes, formally defined as \(\mathcal{D} = \{(x_i, c^{gt}_{i,k}, y_{i,m})\}_{i=1}^N\), where each data point consists of the input \(x_i \in \mathcal{X}\), the label \(y_{i,m} \in \mathcal{Y} \subset \{0, 1\}\), and the concept \(c^{gt}_{i,k} \in \mathcal{C} = \{0, 1\}\). Here, \(k \in \{1, \ldots, K\}\) represents the \(k\)-th concept, and \(m \in \{1, \ldots, M\}\) denotes the \(m\)-th class. A pretrained feature extractor and a concept encoder, jointly represented as a function \(g: \mathcal{X} \to \mathcal{C}\), process the input \(x \in \mathcal{X}\) to predict the concept vector \(\hat{c}_{i,k} \in \mathcal{C}\). Specifically, the feature extractor maps the input \(x_i\) to a latent feature representation \(h_i \in \mathbb{R}^{n_{\text{hidden}}}\), and the concept encoder further transforms \(h_i\) into a set of interpretable concepts \(\hat{c}_{i,k} \in \mathbb{R}^K\). The task classifier, denoted as \(f: \mathbb{R}^K \to \mathcal{Y}\), takes the predicted concepts \(\hat{c}_{i,k}\) as input and maps them to the final class predictions \(\hat{y}_{i,m} \in \mathcal{Y}\). For simplicity, subscript indices are used consistently to denote samples (\(i\)), concepts (\(k\)), and classes (\(m\)).

\subsection{Limitations of CEM}

We consider a Concept Embedding Model (CEM) that processes an input \(x_i \in \mathbb{R}^n\) through a neural network \(\psi: \mathbb{R}^N \to \mathbb{R}^{n_{\text{hidden}}}\), producing a latent representation \(h_i = \psi(x_i)\). Let \(h_i = [r_i, z_i]\), where \(r_i\) contains concept-relevant information and \(z_i\) contains concept-irrelevant features. For each concept \(k \in \{1, \dots, K\}\), the model learns parametric embedding generators \(\phi_k^+(h_i)\) and \(\phi_k^-(h_i)\), which are combined to compute the concept embedding \(\hat{c}_{i,k}\) and its activation probability \(\hat{p}_{i,k}\):
\begin{equation}
    \hat{p}_{i,k} = s\bigl([\phi_k^+(h_i), \phi_k^-(h_i)]\bigr),
    \hat{c}_{i,k} = \hat{p}_{i,k} \hat{c}_{i,k}^+ + (1 - \hat{p}_{i,k}) \hat{c}_{i,k}^-.
\end{equation}

\textbf{Entanglement with Concept-irrelevant Features.}  
The reliability of \(\hat{c}_{i,k}\) relies on its ability to represent concept-relevant information while excluding irrelevant factors. Ideally, \(\hat{c}_{i,k}\) depends solely on \(r_i\). However, due to spurious correlations in the training data, \(\phi_k^\pm(h_i)\) often encodes information from \(z_i\) to minimize the empirical loss:
\begin{equation}
    \min_{\theta} \;\; \mathbb{E}_{(h_i, c^{gt}_{i,k})}\Bigl[\ell\Bigl(s\bigl([\phi_k^+(h_i), \phi_k^-(h_i)]\bigr), c^{gt}_{i,k}\Bigr)\Bigr].
\end{equation}

When \(\hat{c}_{i,k}\) inadvertently incorporates \(z_i\), it becomes sensitive to distributional shifts where \(\Pr(z_i \mid r_i)\) differs from the training set. In such cases, \(\hat{c}_{i,k}\) fails to generalize, and the prediction accuracy of \(\hat{p}_{i,k}\) deteriorates:
\begin{equation}
    \Pr\bigl(\hat{p}_{i,k} \neq c^{gt}_{i,k}\bigr) \text{ increases as irrelevant correlations diminish.}
\end{equation}
This entanglement undermines the reliability of concept embeddings, making them less interpretable and robust.

\textbf{Inconsistency in Concept Semantics.}  
Another critical issue is the inconsistency in the representation of the same concept \(k\) across different instances, even in the absence of concept-irrelevant features \(z_i\). Let two instances have concept-relevant latent representations \(h_{i_1} = r_{i_1}\) and \(h_{i_2} = r_{i_2}\), with \(c^{gt}_{i_1,k} = c^{gt}_{i_2,k}\). Ideally, \(\hat{c}_{i_1,k}\) and \(\hat{c}_{i_2,k}\) should be consistent, i.e., \(\hat{c}_{i_1,k} = \hat{c}_{i_2,k}\). However, due to differences in optimization dynamics or training data distribution, the embedding generators \(\phi_k^\pm(h_i)\) may produce semantic variability, causing \(\hat{c}_{i_1,k} \neq \hat{c}_{i_2,k}\).

To analyze this, consider the expected variability of the concept embedding under a distribution \(\mathcal{D}\) over \(r_i\):
\begin{equation}
    \mathbb{E}_{r_i \sim \mathcal{D}}\Bigl[\|\hat{c}_{i,k} - \bar{c}_k\|_2^2\Bigr],
\end{equation}
where \(\bar{c}_k = \mathbb{E}_{r_i \sim \mathcal{D}}[\hat{c}_{i,k}]\) represents the mean embedding for concept \(k\). A high variance indicates significant inconsistency in the semantic representation of the concept across instances.

Furthermore, this inconsistency can propagate to downstream predictions. Let \(f: \mathbb{R}^{Kd} \to \mathcal{Y}\) denote a downstream task predictor that utilizes the concatenated embeddings \(\bigl[\hat{c}_{i,1}, \dots, \hat{c}_{i,K}\bigr]\). When \(\hat{c}_{i,k}\) is inconsistent, the prediction error increases:
\begin{equation}
    \Pr\bigl(f([\hat{c}_{i,1}, \dots, \hat{c}_{i,K}]) \neq y_i\bigr) > \Pr\bigl(f([\bar{c}_1, \dots, \bar{c}_K]) \neq y_i\bigr),
\end{equation}
where \([\bar{c}_1, \dots, \bar{c}_K]\) represents the consistent mean embeddings. This inconsistency reduces both the interpretability and the robustness of concept-based reasoning.

These two issues—entanglement with irrelevant features and inconsistency in concept representation—highlight the fundamental unreliability of concepts in CEM. Detailed examples and visualizations illustrating these limitations are included in Appendix \ref{appendix:Visualization}.

\subsection{Concept-Level Disentanglement}

To better disentangle concept-irrelevant features, we designed a concept-level disentanglement framework consisting of an encoder-decoder structure.

The encoder extracts concept-irrelevant features from the latent space derived by the Concept Embedding Model (CEM). For an input \(x_i\) with a latent representation \(h_i \in \mathbb{R}^{n_{\text{hidden}}}\), the disentanglement encoder \(E_{\text{dis}}: \mathbb{R}^{n_{\text{hidden}}} \to \mathbb{R}^{Kd}\) produces a disentangled vector \(\hat{z}_{i} \in \mathbb{R}^{Kd}\), separating irrelevant features from the concept embeddings:
\begin{equation}
\hat{z}_{i} = E_{\text{dis}}(h_i). \label{eq:disentangled_vector_combined}
\end{equation}

Concept embeddings \(\tilde{c}_{i,k}\) are defined based on ground-truth concept labels \(c^{gt}_{i,k}\) using a binary mask \(\mu_{i,k} \in \{0, 1\}\). The true embedding is:
\begin{align}
\tilde{c}_{i,k} = \mu_{i,k} \hat{c}_{i,k}^+ + (1 - \mu_{i,k}) \hat{c}_{i,k}^-, \label{eq:true_concept_embedding_combined}
\end{align}
where \(\hat{c}_{i,k}^+\) and \(\hat{c}_{i,k}^-\) represent active and inactive embeddings. The complete embedding for all concepts in sample \(x_i\) is:
\begin{align}
\tilde{\mathbf{C}}_i = \big[\tilde{c}_{i,1}, \ldots, \tilde{c}_{i,K}\big] \in \mathbb{R}^{Kd}.
\end{align}

To enforce disentanglement, adversarial training using a Gradient Reversal Layer (GRL) . The predicted class probabilities are:
\begin{align}
\hat{s}_{i,m} = \frac{\exp(f_m(\mathrm{GRL}(\hat{z}_{i})))}{\sum_{m=1}^M \exp(f_m(\mathrm{GRL}(\hat{z}_{i})))}, \label{eq:softmax_prediction_combined}
\end{align}
where \(f_m\) is the logit corresponding to class \(m\), and \(\mathrm{GRL}\) acts as the identity function during forward propagation but scales the backward gradient by \(-\lambda\). By ensuring \(\hat{z}_{i}\) minimizes its influence on \(\hat{s}_{i}\), classification-related signals are forced into \(\hat{c}_{i}\), leaving \(\hat{z}_{i}\) free of concept-relevant information. To better achieve disentanglement, we use adversarial cross-entropy and Hilbert-Schmidt Independence Criterion (HSIC)\cite{gretton2005measuring} regularization with annealing coefficients as the loss. The concept vector disentanglement loss is defined as:
\begin{align}
\mathcal{L}_{\text{cvd}} = 
-\frac{1}{N} \sum_{i=1}^N \sum_{m=1}^M y_{i,m} \log \hat{s}_{i,m} 
+ \beta \,\mathrm{HSIC}(\hat{z}_{i}, \tilde{\mathbf{C}}_i),
\label{eq:disentanglement_loss_combined}
\end{align}
where \(y_{i,m}\) is the ground-truth label for class \(m\) of sample \(x_i\). \(\beta \in [0,1]\) is a weighting factor that controls the degree of alignment and will increase from 0 during training.

The decoder reconstructs the latent representation \(h_i\) by combining true concept embeddings \(\tilde{\mathbf{C}}_i\) and the disentangled vector \(\hat{z}_{i}\). The decoder \(D: \mathbb{R}^{Kd} \to \mathbb{R}^{n_{\text{hidden}}}\) outputs the reconstructed latent representation:
\begin{align}
\hat{h}_i = D(\tilde{\mathbf{C}}_i, \hat{z}_{i}). \label{eq:decoder_output_combined}
\end{align}

The reconstruction loss ensures the encoder and decoder preserve rich feature information:
\begin{align}
\mathcal{L}_{\text{rec}} = \frac{1}{N} \sum_{i=1}^N \|h_i - \hat{h}_i\|_1. \label{eq:reconstruction_loss_combined}
\end{align}

This process prevents the loss of critical features during disentanglement, ensuring that \(\hat{c}_{i}\) captures concept-relevant signals, and \(\hat{z}_{i}\) isolates irrelevant features.

\subsection{Concept Mixup}
Inconsistency of semantic representations is an important challenge in CBMs. This leads to inconsistent semantic representations of the same concepts in different samples or categories, thus weakening the reliability of concept embedding. In existing approaches~\cite{sheth2024auxiliary,margeloiu2021concept} to enhance CBM by considering concept semantic representation relationships, it is usually assumed that different concepts are independent. Based on this assumption, cosine similarity or orthogonality constraints are usually applied to enhance the independence of different concepts. However, in practical applications, this assumption is often unrealistic. For example, the concept of “blackhead” is usually associated with “blackback” in birds.

Therefore, rather than focusing on ensuring independence between concepts, our approach ensures that semantic representations of the same concepts remain consistent across samples or categories. Specifically, we argue that activation representations of the same concepts should exhibit consistent semantic meanings and should not be significantly influenced by context. By aligning the semantic representations of activated concepts, we can improve the reliability of concept embedding and enhance the performance of downstream tasks.

To address this, we propose a semantic alignment mechanism that ensures consistent representations for the same concept across samples. This mechanism aligns only the embeddings of concepts with true labels \(c^{gt}_{i,k} = 1\), while preserving the original embeddings for inactive concepts.

Let \(\mu_{i,k} \in \{0, 1\}\) be a mask, where \(\mu_{i,k} = 1\) if concept \(k\) is active (\(c^{gt}_{i,k} = 1\)) for sample \(x_i\), and \(\mu_{i,k} = 0\) otherwise. The aligned embedding for concept \(k\) in sample \(x_i\), denoted as \(\dot{c}_{i,k}\), is defined as:
\begin{align}
    \dot{c}_{i,k} = \mu_{i,k} \Big( \beta \bar{c}_k^+ + (1 - \beta) \hat{c}_{i,k}^+ \Big) 
    &+ \big(1 - \mu_{i,k}\big) \hat{c}_{i,k},
    \label{eq:aligned_concept_embedding}
\end{align}
Here, \(\hat{c}_{i,k}^+\) represents the active embedding for concept \(k\) in sample \(x_i\), and \(\bar{c}_k^+\) is the semantic mean embedding of concept \(k\), calculated as:
\begin{align}
    \bar{c}_k^+ = \frac{1}{|S_k|} \sum_{i \in S_k} \hat{c}_{i,k}^+,
    \label{eq:semantic_mean}
\end{align}
where \(S_k = \{i \mid c_{i,k} = 1\}\) is the set of samples in which concept \(k\) is active. For inactive concepts (\(c^{gt}_{i,k} = 0\)), the original embedding \(\hat{c}_{i,k} = \hat{p}_{i,k} \hat{c}_{i,k}^+ + \big(1 - \hat{p}_{i,k}\big) \hat{c}_{i,k}^-\) is used.

The adjusted concept representation for sample \(x_i\) is represented as:
\begin{align}
    \dot{\mathbf{C}}_i = \big[\dot{c}_{i,1}, \dot{c}_{i,2}, \ldots, \dot{c}_{i,K}\big].
    \label{eq:adjusted_representation}
\end{align}

The concept mixup loss considers the adjusted representation \(\dot{\mathbf{C}}_i\) and is defined as:
\begin{align}
    \mathcal{L}_{\text{m}} = -\frac{1}{N} \sum_{i=1}^N \sum_{m=1}^M y_{i,j} \log f_m\Big(
    \big[\dot{c}_{i,1}, \dot{c}_{i,2}, \ldots, \dot{c}_{i,K}\big]\Big),
    \label{eq:classification_loss_adjusted}
\end{align}
where \(N\) is the number of samples, \(M\) is the number of classes, \(y_{i,m}\) is the ground-truth label for class \(m\) of sample \(x_i\), and \(f_m\) is the classifier’s logit for class \(m\). 

The proposed semantic alignment mechanism ensures that the embeddings of the same concept across different samples consistently converge toward a shared semantic mean \(\bar{c}_k^+\). This semantic mean acts as a stable reference point for concept \(k\), derived from multiple active samples, and serves to unify the semantic representation of the concept across diverse contexts. By aligning individual embeddings \(\hat{c}_{i,k}\) with \(\bar{c}_k^+\), the mechanism minimizes intra-concept variability, ensuring that representations of the same concept remain concentrated around a coherent and consistent semantic representation. This semantic alignment enhances the reliability of concept embeddings and strengthens the robustness of the model in downstream tasks.

In summary, the final objective function combines the standard task loss (\(\mathcal{L}_{\text{task}}\)) and concept prediction loss (\(\mathcal{L}_{\text{concept}}\)), proposed in the original CEM framework, with three new losses: concept mixup loss (\(\mathcal{L}_{\text{m}}\)), concept vector disentanglement loss (\(\mathcal{L}_{\text{cvd}}\)), and reconstruction loss (\(\mathcal{L}_{\text{rec}}\)). These losses are combined as follows:
\begin{align}
    \mathcal{L}_{\text{final}} \;=\;& 
    \mathcal{L}_{\text{task}} 
    \;+\; \alpha\,\mathcal{L}_{\text{concept}} \nonumber \\
    &+\; \lambda_{\text{m}}\,\mathcal{L}_{\text{m}} 
    \;+\; \lambda_{\text{cvd}}\,\mathcal{L}_{\text{cvd}} 
    \;+\; \lambda_{\text{rec}}\,\mathcal{L}_{\text{rec}}, \label{eq:final_loss}
\end{align}
where \(\alpha,\lambda_{\text{m}},\lambda_{\text{cvd}},\lambda_{\text{rec}}>0\) are weighting hyperparameters controlling the relative importance of each loss term. A sensitivity analysis for these hyperparameters is presented in Appendix~\ref{appendix:Computation Resources and Training Settings}.

\begin{table*}[!t] 
\centering
\caption{Comparison of Methods and Ablation Study on CUB, CelebA, and AwA2 Datasets (Concept and Task Accuracy Performance)}
\resizebox{\textwidth}{!}{%
\begin{tabular}{lcccccc}
\toprule
\multicolumn{1}{c}{\textbf{Method}} & \multicolumn{2}{c}{\textbf{CUB}} & \multicolumn{2}{c}{\textbf{CelebA}} & \multicolumn{2}{c}{\textbf{AwA2}} \\
\cmidrule(lr){2-3} \cmidrule(lr){4-5} \cmidrule(lr){6-7}
& Concept ($\pm$CI) & Task ($\pm$CI) & Concept ($\pm$CI) & Task ($\pm$CI) & Concept ($\pm$CI) & Task ($\pm$CI) \\
\midrule
Fuzzy-CBM\cite{koh2020concept}    & 95.882 ($\pm$0.105) & 74.228 ($\pm$0.606) & 87.269 ($\pm$0.211) & 33.765 ($\pm$2.158) & 97.000 ($\pm$0.168) & 90.089 ($\pm$1.005) \\
Bool-CBM\cite{koh2020concept}     & 96.229 ($\pm$0.031) & 72.512 ($\pm$0.466) & 86.329 ($\pm$0.164) & 33.952 ($\pm$0.914) & 97.001 ($\pm$0.188) & 89.869 ($\pm$1.047) \\
CEM\cite{zarlenga2022concept}           & 96.160 ($\pm$0.157) & 79.029 ($\pm$0.519) & 87.237 ($\pm$0.306) & 41.719 ($\pm$1.428) & \textbf{98.048 ($\pm$0.037)} & 90.745 ($\pm$0.294) \\
Prob-CBM\cite{kim2023probabilistic}     & 95.596 ($\pm$0.061) & 76.265 ($\pm$0.145) & 87.272 ($\pm$0.238) & 34.487 ($\pm$0.912) & 97.283 ($\pm$0.065) & 90.485 ($\pm$0.315) \\
Coop-CBM\cite{sheth2024auxiliary}     & 89.892 ($\pm$0.649) & 79.054 ($\pm$0.734) & 88.534 ($\pm$0.124) & 40.396 ($\pm$1.365) & 97.875 ($\pm$0.107) & 89.927 ($\pm$0.153) \\
\textbf{RECEM(Ours)} & \textbf{96.560 ($\pm$0.040)} & \textbf{79.831 ($\pm$0.402)} & \textbf{88.617 ($\pm$0.309)} & \textbf{50.143 ($\pm$0.860)} & 97.849 ($\pm$0.022) & \textbf{91.363 ($\pm$0.102)} \\
\midrule
\textbf{w/o $\mathcal{L}_{\text{m}}$} & 96.112 ($\pm$0.125) & 79.165 ($\pm$0.467) & 88.283 ($\pm$0.291) & 45.295 ($\pm$0.822) & 97.750 ($\pm$0.031) & 90.712 ($\pm$0.144) \\
\textbf{w/o $\mathcal{L}_{\text{rec}}$} & 96.393 ($\pm$0.002) & 79.463 ($\pm$0.452) & 87.553 ($\pm$0.099) & 46.917 ($\pm$0.114) & 97.700 ($\pm$0.025) & 90.540 ($\pm$0.121) \\
\textbf{w/o $\mathcal{L}_{\text{cvd}}$} & 96.471 ($\pm$0.012) & 79.572 ($\pm$0.483) & 87.153 ($\pm$1.125) & 46.783 ($\pm$1.820) & 97.802 ($\pm$0.033) & 90.622 ($\pm$0.130) \\
\textbf{w/o $\mathcal{L}_{\text{cvd}}, \mathcal{L}_{\text{rec}}$} & 96.388 ($\pm$0.142) & 79.498 ($\pm$0.510) & 87.322 ($\pm$0.161) & 46.746 ($\pm$0.433) & 97.652 ($\pm$0.048) & 90.552 ($\pm$0.144) \\
\textbf{w/o $\mathcal{L}_{\text{cvd}}, \mathcal{L}_{\text{m}}$} & 96.115 ($\pm$0.125) & 79.162 ($\pm$0.467) & 88.281 ($\pm$0.291) & 45.292 ($\pm$0.822) & 97.601 ($\pm$0.051) & 89.902 ($\pm$0.157) \\
\textbf{w/o $\mathcal{L}_{\text{rec}}, \mathcal{L}_{\text{m}}$} & 96.350 ($\pm$0.137) & 78.893 ($\pm$0.452) & 88.550 ($\pm$0.299) & 43.517 ($\pm$0.714) & 97.721 ($\pm$0.035) & 90.003 ($\pm$0.119) \\
\bottomrule
\end{tabular}%
}
\label{tab:comparison_and_ablation}
\end{table*}


\section{Experiments}
\subsection{Experiment Setup}
\textbf{Datasets} We evaluated RECEM on four datasets: CUB, TravelingBirds, CelebA, and AwA2. The CUB dataset \cite{welinder2010caltech} is a bird image dataset with 112 annotated attributes serving as concepts, following prior work \cite{koh2020concept, zarlenga2022concept}. TravelingBirds \cite{koh2020concept} is a modified version of CUB where image backgrounds are replaced using the Places dataset \cite{7968387}. The CelebA dataset \cite{liu2015deep} contains facial images with 40 binary attributes used as concepts, while the AwA2 dataset \cite{xian2018zero} focuses on animal classes with 85 annotated attributes serving as concepts. Additional details can be found in the Appendix~\ref{appendix:Detailed Dataset Descriptions}.

\textbf{Baselines and Implementation Details} RECEM was compared with CBM \cite{koh2020concept}, ProbCBM \cite{kim2023probabilistic}, Coop-CBM \cite{sheth2024auxiliary}, and CEM \cite{zarlenga2022concept}, all using ResNet34 \cite{he2016deep} as the feature extractor. Models were trained with identical settings using SGD as the optimizer. For RECEM, the concept mixup loss weight was \(\lambda_{\text{m}} = 0.1\), the concept vector disentanglement loss weight was \(\lambda_{\text{cvd}} = 0.05\), and the reconstruction loss weight was \(\lambda_{\text{rec}} = 1\), while other settings followed CEM defaults. All baseline models were independently trained by us using the original implementations provided by their authors. Results were averaged over five random seeds. Training configurations and computational details are in the Appendix~\ref{appendix:Computation Resources and Training Settings}.

\subsection{Concept and Task Performance}

We evaluate RECEM on the CUB, CelebA, and AwA2 datasets, comparing its performance in concept prediction, downstream task accuracy, and Concept Alignment Score (CAS)~\cite{zarlenga2022concept} with baseline methods, as shown in Tables~\ref{tab:comparison_and_ablation} and~\ref{tab:cas}. CAS evaluates the alignment of concepts and their ground truth labels, where higher scores indicate more reliable concept representations.

On the CUB dataset, RECEM achieves the highest task accuracy of 79.831\%, surpassing all baselines, including CEM (79.029\%) and Coop-CBM (79.054\%). RECEM also achieves the highest concept accuracy of 96.560\% and a CAS of 93.206, significantly outperforming CEM's CAS of 86.143. These results validate the effectiveness of the Concept Mixup mechanism and disentanglement strategy in reducing the influence of irrelevant features and ensuring consistent, semantically aligned concept representations.

Similarly, on the CelebA dataset, RECEM achieves substantial improvements in task accuracy (50.143\% vs. Coop-CBM's 40.396\% and CEM's 41.719\%) and CAS (84.367 vs. CEM's 79.471). These results highlight the robustness of RECEM in real-world scenarios with diverse visual variations, ensuring both reliable concept representations and strong downstream performance.

On the AwA2 dataset, RECEM achieves the best task accuracy of 91.363\% and the highest CAS of 95.653, surpassing all baselines. Although the concept accuracy (97.849\%) is slightly lower than CEM’s 98.048\%, the improved CAS demonstrates that RECEM provides more faithful concept representations while maintaining robust downstream predictions.

\begin{table}[t]
\centering
\caption{Comparison of Concept Alignment Score (CAS) across different datasets. Higher CAS values indicate better performance.}
\resizebox{\columnwidth}{!}{%
\begin{tabular}{lccc}
\hline
\textbf{} & \textbf{CUB} & \textbf{CelebA} & \textbf{AwA2} \\
\hline
CEM & 86.143 $\pm$ 0.631 & 79.471 $\pm$ 0.985 & 94.893 $\pm$ 0.429 \\
Fuzzy-CBM & 80.093 $\pm$ 0.941 & 74.990 $\pm$ 0.351 & 88.863 $\pm$ 0.342 \\
\textbf{RECEM (Ours)} & \textbf{93.206 $\pm$ 0.488} & \textbf{84.367 $\pm$ 0.493} & \textbf{95.653 $\pm$ 0.229} \\
\hline
\end{tabular}%
}
\label{tab:cas}
\end{table}

\textbf{Ablation studies.} Ablation studies demonstrate the importance of the key components in RECEM: Concept Mixup loss (\(\mathcal{L}_{\text{m}}\)), concept vector disentanglement loss (\(\mathcal{L}_{\text{cvd}}\)), and reconstruction loss (\(\mathcal{L}_{\text{rec}}\)). Removing \(\mathcal{L}_{\text{m}}\) reduces task accuracy significantly, particularly on CUB (from 79.831\% to 79.165\%), highlighting its role in maintaining consistent concept representations. Excluding \(\mathcal{L}_{\text{cvd}}\), which disentangles concept vectors via adversarial learning, causes notable performance drops, especially on CelebA (from 50.143\% to 46.783\%), demonstrating its critical role in robust concept representations. Similarly, removing \(\mathcal{L}_{\text{rec}}\) leads to consistent declines, confirming its importance in preserving rich semantic details.

\begin{figure}[!t]
    \centering
    \includegraphics[trim=0 475pt 0 0, clip, width=\columnwidth]{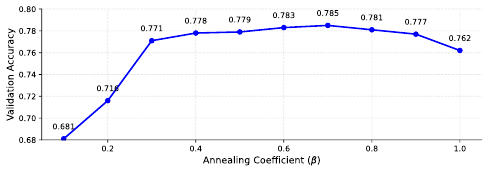}
    \caption{The impact of the annealing coefficient \(\beta\)}
    \label{fig:beta_analysis}
\end{figure}

We further analyze the impact of the annealing coefficient \(\beta\), which facilitates better convergence by gradually increasing the strength of semantic alignment and HSIC regularization. As shown in Figure~\ref{fig:beta_analysis}, increasing \(\beta\) initially improves validation accuracy, reflecting the benefits of stronger alignment and disentanglement.However, when $\beta$ becomes too large, the accuracy slightly declines, possibly because an excessively large $\beta$ interferes with the model's effective convergence.

Compared to all baselines, RECEM consistently achieves superior task accuracy. These results underscore the importance of the disentanglement mechanism and the Concept Mixup mechanism in enhancing both concept prediction and task performance, establishing RECEM as a reliable and robust framework for interpretable machine learning.

\begin{figure*}[!t] 
    \centering
    \includegraphics[trim=0 390pt 0 0, clip,width=\textwidth]{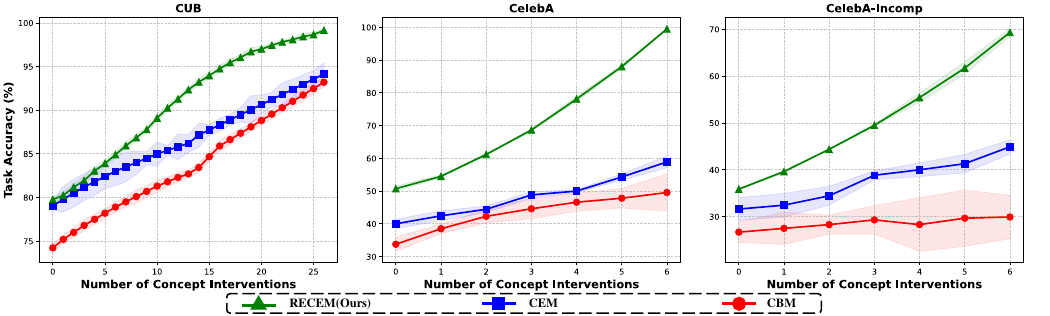}
    \caption{Task accuracy under varying levels of human intervention in concept predictions for different models.}
    \label{fig:intervention}
\end{figure*} 

\subsection{Concept Interventions}

To evaluate the robustness and reliability of our model, we conducted a series of concept intervention experiments. These experiments were designed to analyze how human intervention, aimed at correcting concept predictions, influences overall task accuracy. We conducted experiments on the CUB, CelebA and CelebA-Incomp datasets, where CelebA-Incomp is a conceptually incomplete version of CelebA that follows the CEM's setting\cite{zarlenga2022concept}, aiming to evaluate the behaviour of each method in scenarios where conceptual annotations are sparse and incomplete.

As illustrated in the figure~\ref{fig:intervention}, for the CUB and CelebA datasets, RECEM consistently outperformed the baseline models as the intervention ratio increased, achieving near 100\% task accuracy at higher intervention ratios. This demonstrates that reliable concept representation enables the model to respond more effectively to human interventions. On the CelebA-Incomp dataset, our model also showed substantial improvements in task accuracy with increasing intervention ratios, consistently outperforming all baseline methods. This highlights the superior performance and robustness of our method even when concept annotations are incomplete.

Effectively leveraging human interventions is a critical aspect of the interpretability of concept bottleneck models. The inability of baseline methods to efficiently utilize human interventions underscores how the interference of concept-irrelevant information in concept encoding undermines their interpretability.

\subsection{Robustness to Background Shifts}

\begin{table}[t]
\centering
\caption{Task Accuracy with background shifts.}
\resizebox{\columnwidth}{!}{%
\begin{tabular}{lccc}
\hline
\textbf{} & \textbf{CUB\_Random} & \textbf{CUB\_Fixed/test} & \textbf{CUB\_Black} \\
\hline
ProbCBM & 60.231 $\pm$ 0.754 & 60.089 $\pm$ 0.801 & 61.104 $\pm$ 2.421 \\
CEM & 60.198 $\pm$ 0.395 & 60.034 $\pm$ 0.416 & 62.112 $\pm$ 3.279 \\
Fuzzy-CBM & 59.003 $\pm$ 0.511 & 59.100 $\pm$ 0.487 & 59.080 $\pm$ 0.502 \\
Bool-CBM & 56.200 $\pm$ 1.743 & 56.145 $\pm$ 1.811 & 56.089 $\pm$ 1.792 \\
Coop-CBM & 59.478 $\pm$ 0.945 & 59.132 $\pm$ 1.142 & 61.314 $\pm$ 0.231 \\
\textbf{RECEM (Ours)} & \textbf{64.102 $\pm$ 1.412} & \textbf{63.987 $\pm$ 1.533} & \textbf{65.003 $\pm$ 1.467} \\
\hline
\end{tabular}%
}
\label{tab:bg_task_accuracy}
\end{table}

To evaluate the robustness of RECEM to background variations, we conduct experiments on the TravelingBirds dataset, a variant of the CUB dataset designed with systematic background shifts~\cite{koh2020concept}. RECEM is trained on the original CUB dataset and tested on three variations: CUB\_Random, CUB\_Fixed/test, and CUB\_Black. These variations simulate real-world scenarios, challenging the model's ability to focus on concept-relevant features.

As shown in Table~\ref{tab:bg_task_accuracy}, RECEM achieves the highest accuracy across all three datasets, consistently outperforming baseline methods. In scenarios like CUB\_Random and CUB\_Fixed/test, where background changes introduce significant distribution shifts, RECEM demonstrates superior robustness, maintaining stable performance compared to notable drops in accuracy observed in baselines. This indicates that RECEM effectively prioritizes concept-relevant features while mitigating interference from background noise.

The strong performance of RECEM can be attributed to its Concept Mixup mechanism and Concept-Level Disentanglement framework. The Concept Mixup mechanism enforces semantic alignment of concept embeddings, ensuring that representations are consistent and robust across samples despite variations in background. Meanwhile, the disentanglement framework isolates irrelevant background features, allowing downstream predictions to rely exclusively on object-centric concepts. Together, these mechanisms enable RECEM to maintain reliable performance in the presence of diverse background conditions, demonstrating its suitability for real-world tasks where background variations are prevalent.

\section{Conclusion}

We proposed the Reliability-Enhanced Concept Embedding Model (RECEM), a framework designed to tackle challenges of concept reliability and robustness in Concept Bottleneck Models. By integrating Concept-Level Disentanglement to isolate irrelevant features and Concept Mixup to ensure semantic alignment, RECEM achieves reliable and robust concept representations. Experiments across diverse datasets, including scenarios with significant background and domain shifts, demonstrate consistent improvements in performance, validating RECEM's ability to generate reliable concept representations.


\bibliography{ref}
\bibliographystyle{arxiv}

\newpage
\appendix
\onecolumn
\section{Appendix}
\subsection{Detailed Dataset Descriptions}\label{appendix:Detailed Dataset Descriptions}

\textbf{CUB} The Caltech-UCSD Birds-200-2011 (CUB) dataset~\cite{welinder2010caltech} is a benchmark for fine-grained classification and contains 200 bird species with annotated attributes. Following prior work~\cite{koh2020concept, zarlenga2022concept}, we use 112 binary bird attributes (e.g., \textit{beak\_type}, \textit{wing\_color}) as concepts and the bird identity (\(m = 200\)) as the downstream task label. The dataset is split into training, validation, and testing sets as defined in~\cite{koh2020concept}.

\textbf{TravelingBirds} TravelingBirds~\cite{koh2020concept} is a variant of the CUB dataset, where the original background of each image is replaced with an image from the Places dataset~\cite{7968387}. This setup introduces systematic background shifts to evaluate model robustness under domain changes. The same 112 binary attributes and bird identity are used as the downstream task label.

\textbf{CelebA and CelebA-incomp} The CelebFaces Attributes (CelebA) dataset~\cite{liu2015deep} contains facial images annotated with 40 binary attributes. In our experiments, we use two configurations of this dataset. The first, CelebA, uses 6 attributes (\(a_1, \dots, a_6\)) as binary concept annotations, with the downstream task predicted using the base-10 representation of these attributes, resulting in \(l = 2^6 = 64\) classes. The dataset is split following the protocol in CEM~\cite{zarlenga2022concept}. The second configuration, CelebA-incomp, simulates incomplete concept annotations by excluding \(a_7\) and \(a_8\), which are critical for task prediction. While the remaining attributes (\(a_1, \dots, a_6\)) are used as binary concepts, the downstream task still relies on all 8 attributes, evaluating the model's performance when concept annotations are sparse or incomplete.

\textbf{AwA2} The Animals with Attributes 2 (AwA2) dataset~\cite{xian2018zero} comprises 50 animal classes annotated with 85 attributes.It comprises images of animals representing 50 different species, totaling 37 322 images. Each image is annotated with 85 attributes delineating various animal characteristics, including color, habitat, and diet. These attributes are used as binary concept annotations, while the animal classes serve as downstream task labels.

\subsection{Computation Resources and Training Settings}\label{appendix:Computation Resources and Training Settings}

\begin{table*}[h]
\centering
\caption{Hyperparameter Sensitivity Analysis for \(\lambda_{\text{m}}\), \(\lambda_{\text{cvd}}\), and \(\lambda_{\text{rec}}\) on CUB, CelebA, and AwA2 Datasets.}
\resizebox{\textwidth}{!}{%
\begin{tabular}{lcccccc}
\hline
\textbf{Hyperparameter} & \multicolumn{2}{c}{\textbf{CUB}} & \multicolumn{2}{c}{\textbf{CelebA}} & \multicolumn{2}{c}{\textbf{AwA2}} \\
\cmidrule(lr){2-3} \cmidrule(lr){4-5} \cmidrule(lr){6-7}
& Concept ($\pm$CI) & Task ($\pm$CI) & Concept ($\pm$CI) & Task ($\pm$CI) & Concept ($\pm$CI) & Task ($\pm$CI) \\
\hline
\textbf{\(\lambda_{\text{m}} = 0.01\)} & 96.142 ($\pm$0.153) & 79.231 ($\pm$0.411) & 88.432 ($\pm$0.299) & 49.621 ($\pm$0.252) & 97.702 ($\pm$0.029) & 90.714 ($\pm$0.107) \\
\textbf{\(\lambda_{\text{m}} = 0.1\)} & \textbf{96.560 ($\pm$0.143)} & \textbf{79.831 ($\pm$0.402)} & \textbf{88.617 ($\pm$0.309)} & \textbf{50.143 ($\pm$0.860)} & \textbf{97.849 ($\pm$0.037)} & \textbf{91.363 ($\pm$0.102)} \\
\textbf{\(\lambda_{\text{m}} = 0.5\)} & 96.032 ($\pm$0.165) & 78.742 ($\pm$0.442) & 88.098 ($\pm$0.321) & 49.143 ($\pm$0.278) & 97.561 ($\pm$0.035) & 90.213 ($\pm$0.115) \\
\textbf{\(\lambda_{\text{m}} = 1.0\)} & 95.892 ($\pm$0.172) & 78.342 ($\pm$0.463) & 87.872 ($\pm$0.325) & 48.763 ($\pm$0.213) & 97.503 ($\pm$0.038) & 89.872 ($\pm$0.122) \\
\hline
\textbf{\(\lambda_{\text{cvd}} = 0.01\)} & 96.324 ($\pm$0.148) & 79.432 ($\pm$0.425) & 88.521 ($\pm$0.309) & 49.798 ($\pm$0.249) & 97.789 ($\pm$0.028) & 90.943 ($\pm$0.104) \\
\textbf{\(\lambda_{\text{cvd}} = 0.05\)} & \textbf{96.560 ($\pm$0.143)} & \textbf{79.831 ($\pm$0.402)} & \textbf{88.617 ($\pm$0.309)} & \textbf{50.143 ($\pm$0.860)} & \textbf{97.849 ($\pm$0.037)} & \textbf{91.363 ($\pm$0.102)} \\
\textbf{\(\lambda_{\text{cvd}} = 0.1\)} & 96.132 ($\pm$0.162) & 78.972 ($\pm$0.452) & 88.248 ($\pm$0.318) & 49.401 ($\pm$0.287) & 97.692 ($\pm$0.032) & 90.714 ($\pm$0.113) \\
\textbf{\(\lambda_{\text{cvd}} = 0.5\)} & 95.924 ($\pm$0.173) & 78.471 ($\pm$0.465) & 88.032 ($\pm$0.323) & 49.093 ($\pm$0.215) & 97.607 ($\pm$0.036) & 89.498 ($\pm$0.118) \\
\hline
\textbf{\(\lambda_{\text{rec}} = 0.1\)} & \textbf{96.634 ($\pm$0.152)} & 79.562 ($\pm$0.435) & \textbf{88.735 ($\pm$0.298)} & 49.842 ($\pm$0.718) & 97.837 ($\pm$0.031) & 90.881 ($\pm$0.109) \\
\textbf{\(\lambda_{\text{rec}} = 0.5\)} & 96.411 ($\pm$0.147) & 79.671 ($\pm$0.421) & 88.598 ($\pm$0.306) & 50.124 ($\pm$0.712) & 97.842 ($\pm$0.027) & 90.952 ($\pm$0.106) \\
\textbf{\(\lambda_{\text{rec}} = 1.0\)} & 96.560 ($\pm$0.143) & \textbf{79.831 ($\pm$0.402)} & 88.617 ($\pm$0.309) & \textbf{50.143 ($\pm$0.860)} & \textbf{97.849 ($\pm$0.037)} & \textbf{91.363 ($\pm$0.102)} \\
\textbf{\(\lambda_{\text{rec}} = 5.0\)} & 96.210 ($\pm$0.159) & 79.509 ($\pm$0.432) & 88.398 ($\pm$0.318) & 49.763 ($\pm$0.281) & 97.821 ($\pm$0.033) & 90.783 ($\pm$0.112) \\
\hline
\end{tabular}%
}
\label{tab:combined_sensitivity}
\end{table*}

\textbf{Computation Resources} Our experiments were conducted using different GPU clusters. Baseline experiments utilized NVIDIA A100 GPUs for training, while ablation studies and background shift experiments were performed on shared clusters with NVIDIA A5000 GPUs.

\textbf{Hyperparameter Settings} The key hyperparameters for our method include the concept mixup loss weight \(\lambda_{\text{m}}\), the concept vector disentanglement loss weight \(\lambda_{\text{cvd}}\), and the reconstruction loss weight \(\lambda_{\text{rec}}\). By default, these were set to \(\lambda_{\text{m}} = 0.1\), \(\lambda_{\text{cvd}} = 0.05\), and \(\lambda_{\text{rec}} = 1.0\), as they provided the best performance in preliminary experiments. For fairness, all models were trained under the same optimization settings using SGD, and results were averaged across five random seeds.

\textbf{Model Architectures} To ensure a fair comparison between methods, we used identical architectures for the concept encoder \(g\) and label predictor \(f\) across all baselines, following the architecture choices outlined in CEM~\cite{zarlenga2022concept}. For both the encoder and decoder in our concept-level disentanglement module, we implemented single linear layers to align with the fairness and simplicity of architecture choices. Additionally, we followed CEM's settings for concept embedding dimensions (\(d=16\)) and used the same Randint value of 0.25.

\textbf{Sensitivity Analysis} To evaluate the impact of these hyperparameters, we performed sensitivity analyses on \(\lambda_{\text{m}}\), \(\lambda_{\text{cvd}}\), and \(\lambda_{\text{rec}}\), as summarized in Table~\ref{tab:combined_sensitivity}.
For \(\lambda_{\text{cvd}}\), smaller values fail to disentangle concept vector-level features adequately, leading to less robust representations. Larger values, however, overfit the adversarial loss, distorting task-relevant features. For \(\lambda_{\text{rec}}\), smaller values hinder the ability to reconstruct rich latent features, while larger values overemphasize reconstruction at the expense of disentanglement. Similarly, for \(\lambda_{\text{m}}\), smaller values reduce semantic alignment across concepts, while larger values overly constrain the concept encoder.

\subsection{Visualization}\label{appendix:Visualization}

To provide a more intuitive understanding of how concept-irrelevant features affect concept encoding and downstream task performance, we include visualizations in this section. These visualizations aim to highlight the influence of irrelevant features on concept representations and demonstrate the robustness of our proposed disentanglement approach.

\begin{figure}[h]
    \centering
    \includegraphics[trim=0 480pt 0 0, clip, width=\columnwidth]{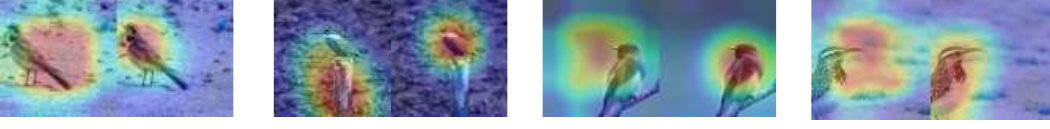}
    \caption{Each pair of heatmaps compares the attention focus of the undecoupled concept encoder (left) and the disentangled concept encoder (right). The images clearly show how our method effectively isolates concept-relevant features, avoiding spurious background correlations.}
    \label{fig:heatmaps}
\end{figure}

Specifically, we present heatmaps~\ref{fig:heatmaps} that illustrate the activation locations of concept encoders under different scenarios, including cases where spurious correlations exist between concepts and background features. Each pair of heatmaps includes the attention focus of an undecoupled concept encoder (left) and that of a disentangled concept encoder (right), offering a clear comparison of how our method successfully isolates concept-relevant features. 

\begin{figure}[h]
    \centering
    \includegraphics[trim=0 445pt 0 0, clip, width=\columnwidth]{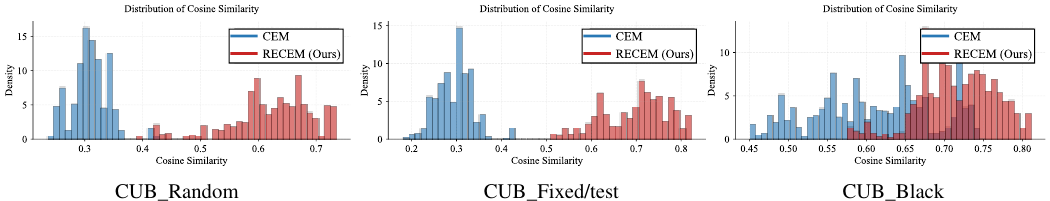}
    \caption{Cosine similarity distributions of concept embeddings before and after background variations. RECEM maintains higher similarity, showing its robustness against background shifts compared to baseline methods.}
    \label{fig:vis_dis}
\end{figure}

In addition to heatmaps, Figure~\ref{fig:vis_dis} visualizes the distribution similarity of concept embeddings under background variations. By analyzing cosine similarity before and after systematic background changes, we evaluate how different models handle distribution shifts. It can be seen that the CEM exhibits worse embedding representation consistency than the RECEM on all the different datasets. The results demonstrate that our RECEM model significantly reduces the impact of background features on concept embeddings, ensuring robust and consistent representations.

\begin{figure}[h]
    \centering
    \includegraphics[trim=0 280pt 0 0, clip, width=\columnwidth]{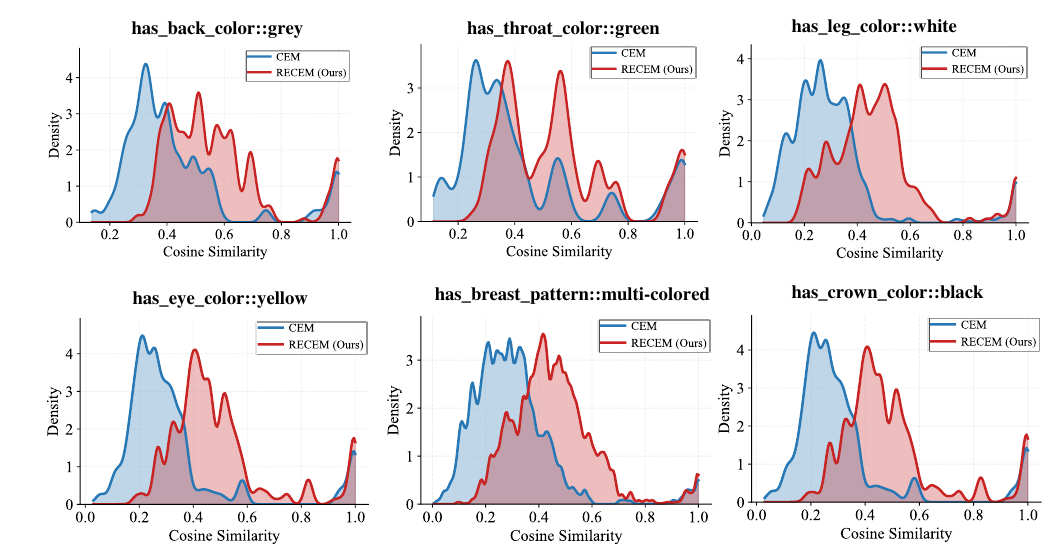}
    \caption{Cosine similarity distributions of concept embeddings before and after background variations. RECEM maintains higher similarity, showing its robustness against background shifts compared to baseline methods.}
    \label{fig:vis_con}
\end{figure}
To provide further evidence of semantic inconsistency, we include visualizations in Figure~\ref{fig:vis_con} that depict the concept representations. Additionally, we randomly selected six more concepts from the CUB dataset and analyzed their cosine similarity distributions. The visualizations and results highlight consistent variability across samples for these concepts, emphasizing the widespread nature of the issue. These examples complement the analysis in the main text, offering a clearer understanding of the challenges faced by traditional CBMs in ensuring semantic consistency.

\subsection{Concept Leakage Within RECEM}\label{appendix:Concept Leakage Within RECEM}

Concept Bottleneck Models often encode irrelevant features within their conceptual representations, leading to a phenomenon known as concept leakage~\cite{mahinpei2021promises,havasi2022addressing}. This diminishes both the interpretability of concepts and the effectiveness of conceptual interventions. To quantify and analyze concept leakage in RECEM, we employ two complementary metrics: Oracle Impurity Score (OIS)~\cite{mahinpei2021promises} and Concept Alignment Score (CAS)~\cite{zarlenga2022concept}. OIS measures the degree of irrelevant information encoded within concepts, where lower scores indicate reduced leakage. CAS evaluates the alignment between learned concepts and their ground truth labels, where higher scores indicate more reliable concept representations.

We conduct experiments on the CUB and CelebA datasets to evaluate concept leakage and fidelity. The CUB dataset includes semantically correlated concepts (e.g., "has\_back\_color::black" and "has\_underparts\_color::black"), while the CelebA dataset contains largely independent concepts. This distinction allows us to investigate the effect of concept correlations on OIS and CAS metrics.

Table~\ref{tab:ois_cas_combined} presents the results for RECEM and the baseline CEM. On the CUB dataset, RECEM achieves a higher CAS (\(93.206 \pm 0.488\)) than CEM (\(86.143 \pm 0.631\)), reflecting the effectiveness of RECEM in ensuring reliable concept representations. However, RECEM reports a higher OIS (\(56.541 \pm 0.30\)) compared to CEM (\(43.54 \pm 2.30\)), which we attribute to the semantic correlations among concepts in the CUB dataset. These correlations conflict with the independence assumption underlying the OIS metric, leading to an inflated score for RECEM.

Conversely, on the CelebA dataset, where concepts are relatively independent, RECEM achieves both a lower OIS (\(27.962 \pm 8.192\)) and a higher CAS (\(84.367 \pm 0.493\)) than CEM (\(40.962 \pm 4.192\) and \(79.471 \pm 0.985\), respectively). These results highlight the robustness of RECEM in reducing irrelevant information and ensuring faithful concept representations, particularly in scenarios with semantically independent concepts.

While OIS provides insights into the degree of irrelevant information encoded within concepts, its reliance on the assumption of semantic independence may limit its applicability in datasets with inherently correlated concepts. In contrast, CAS more directly reflects the reliability of learned concepts, offering a complementary perspective. These findings suggest that while RECEM effectively enhances the reliability of concept representations (as evidenced by CAS), it may not always achieve optimal OIS due to dataset-specific characteristics. This observation also highlights the limitations of OIS as a standalone metric for measuring concept leakage. Future research could explore more comprehensive metrics to better quantify and assess concept leakage in diverse datasets.
\begin{table*}[!t]
\centering
\caption{Oracle Impurity Score (OIS $\downarrow$) and Concept Alignment Score (CAS $\uparrow$) for CEM and RECEM on CUB and CelebA Datasets. OIS measures concept leakage, where lower scores are better, while CAS measures how much learnt concept representations can be trusted as faithful representations of their ground truth concept labels, where higher scores are better.}
\begin{tabular}{lcccc}
\toprule
\textbf{Method} & \multicolumn{2}{c}{\textbf{CUB}} & \multicolumn{2}{c}{\textbf{CelebA}} \\
\cmidrule(lr){2-3} \cmidrule(lr){4-5}
& \textbf{OIS ($\pm$CI) $\downarrow$} & \textbf{CAS ($\pm$CI) $\uparrow$} & \textbf{OIS ($\pm$CI) $\downarrow$} & \textbf{CAS ($\pm$CI) $\uparrow$} \\
\midrule
CEM & \(\textbf{43.54} \pm 2.30\) & \(86.143 \pm 0.631\) & \(40.962 \pm 4.192\) & \(79.471 \pm 0.985\) \\
RECEM & \(56.541 \pm 0.30\) & \(\textbf{93.206} \pm 0.488\) & \(\textbf{27.962} \pm 8.192\) & \(\textbf{84.367} \pm 0.493\) \\
\bottomrule
\end{tabular}
\label{tab:ois_cas_combined}
\end{table*}

\subsection{Extending to Traditional CBMs}\label{appendix:Extending to Traditional CBMs}

To incorporate both concept disentanglement and semantic alignment within the framework of traditional Concept Bottleneck Models, it is necessary to use activation-based representations of concepts. However, traditional CBMs rely on predefined, fixed values for bottleneck activations, such as binary values (e.g., 1 for active and 0 for inactive) or percentile thresholds (e.g., the 95th percentile of activation). This rigid structure effectively blocks the flow of gradients from the label predictor to the concept encoder, which poses a significant challenge to the integration of our proposed framework into traditional CBMs.

Additionally, traditional CBMs lack explicit semantic information regarding the activation and deactivation of concepts. This architectural limitation results in a mismatch between the framework and the training paradigm, undermining the interpretability of traditional CBMs when combined with our mechanisms.

In light of these challenges, we explored an alternative approach that modifies the architecture and sacrifices partial interpretability for compatibility with traditional CBMs. Instead of relying on fixed real concept representations, we replaced them with activation-based representations derived from the concept embedding space. This adjustment facilitates the incorporation of our mechanism into the traditional CBM framework. While this modification deviates from the original design philosophy of traditional CBMs, it allows the mechanisms process to function effectively.

\begin{table*}[!t]
\centering
\caption{Comparison of Bool-CBM and Fuzzy-CBM with and without Concept-level Disentanglement and Concept Mixup on CUB, CelebA, and AwA2 Datasets (Concept and Task Performance).}
\resizebox{\textwidth}{!}{%
\begin{tabular}{lcccccc}
\toprule
\textbf{Method} & \multicolumn{2}{c}{\textbf{CUB}} & \multicolumn{2}{c}{\textbf{CelebA}} & \multicolumn{2}{c}{\textbf{AwA2}} \\
\cmidrule(lr){2-3} \cmidrule(lr){4-5} \cmidrule(lr){6-7}
& Concept ($\pm$CI) & Task ($\pm$CI) & Concept ($\pm$CI) & Task ($\pm$CI) & Concept ($\pm$CI) & Task ($\pm$CI) \\
\midrule
Bool-CBM (original)      & 96.229 ($\pm$0.031) & 73.512 ($\pm$0.466) & 86.329 ($\pm$0.164) & 33.952 ($\pm$0.914) & 97.001 ($\pm$0.188) & 89.869 ($\pm$1.047) \\
Bool-CBM (with mechanisms) & 96.439 ($\pm$0.024) & 73.855 ($\pm$0.421) & 87.489 ($\pm$0.151) & 34.002 ($\pm$3.827) & 97.152 ($\pm$0.173) & 89.870 ($\pm$1.038) \\
Fuzzy-CBM (original)     & 95.882 ($\pm$0.105) & 74.228 ($\pm$0.606) & 87.269 ($\pm$0.211) & 33.765 ($\pm$2.158) & 97.000 ($\pm$0.168) & 90.089 ($\pm$1.005) \\
Fuzzy-CBM (with mechanisms) & 96.135 ($\pm$0.087) & 77.929 ($\pm$0.493) & 87.782 ($\pm$0.193) & 44.301 ($\pm$1.874) & 97.281 ($\pm$0.142) & 90.741 ($\pm$0.902) \\
\bottomrule
\end{tabular}%
}
\label{tab:bool_fuzzy_comparison}
\end{table*}

The results shown in Table \ref{tab:bool_fuzzy_comparison} demonstrate the impact of incorporating both the concept disentanglement mechanism and the semantic alignment mechanism into Bool-CBM and Fuzzy-CBM across the CUB, CelebA, and AwA2 datasets. The comparison highlights several notable trends.

For Bool-CBM, the inclusion of these mechanisms led to slight improvements in both concept and task accuracy on the CUB and CelebA datasets. Specifically, the concept accuracy increased from \(96.229 \%\) to \(96.439 \%\) on the CUB dataset and from \(86.329 \%\) to \(87.489 \%\) on CelebA. However, task accuracy showed only marginal improvement, with no statistically significant changes observed across datasets. This indicates that while concept disentanglement and semantic alignment enhance the representation of concepts, the rigid nature of Bool-CBM's predefined activations limits its ability to fully leverage these refined features for task prediction.

For Fuzzy-CBM, the integration of both mechanisms had a more pronounced effect. On the CUB dataset, task accuracy improved significantly from \(74.228 \%\) to \(77.929 \%\), approaching the performance of more advanced baselines like CEM. A similar trend was observed on CelebA, where task accuracy increased by over 10 percentage points (from \(33.765 \%\) to \(44.301 \%\)). These improvements suggest that Fuzzy-CBM benefits more from disentanglement and alignment, which allow it to better incorporate refined concept representations.

Across all datasets, concept accuracy improved modestly for both models, reflecting the combined effectiveness of disentanglement and semantic alignment in mitigating the influence of irrelevant information. Notably, the task accuracy gains for Fuzzy-CBM highlight the value of these mechanisms in enabling models to more effectively utilize concept-relevant information for robust task predictions.


\end{document}